\documentclass[twoside,leqno,twocolumn]{article}

\usepackage[letterpaper]{geometry}
\usepackage{amsmath}

\usepackage{ltexpprt}
\usepackage{hyperref}
\usepackage{color,soul}
\usepackage{float}
\providecommand{\keywords}[1]
{
  \small	
  \textbf{\textit{Keywords---}} #1
}

\usepackage{amsmath}
\usepackage{graphicx}

\begin{document}

\title{\Large A Deep Adversarial Model for Suffix and Remaining Time Prediction of\\Event Sequences 

}
\author{Farbod Taymouri \thanks{School of Computing and Information Systems, The University of Melbourne, Melbourne, Australia. E-mail: \{farbod.taymouri, marcello.larosa, sarah.erfani\}@unimelb.edu.au}
\and Marcello La Rosa \footnotemark[1]
\and Sarah M. Erfani \footnotemark[1]}

\date{}

\maketitle


\fancyfoot[R]{\scriptsize{Copyright \textcopyright\ 2021 by SIAM\\
Unauthorized reproduction of this article is prohibited}}





\begin{abstract} \small\baselineskip=12pt 

Event suffix and remaining time prediction are sequence to sequence learning tasks. 
They have wide applications in different areas such as economics, digital health, business process management and IT infrastructure monitoring. Timestamped event sequences contain ordered events which carry at least two attributes: the event's label and its timestamp. 
Suffix and remaining time prediction are about obtaining the most likely continuation of event labels and the remaining time until the sequence finishes, respectively.
Recent deep learning-based works for such predictions are prone to potentially large prediction errors because of closed-loop training  (i.e., the next event is conditioned on the ground truth of previous events) and open-loop inference  (i.e., the next event is conditioned on previously predicted events).
In this work, we propose an encoder-decoder architecture for open-loop training to advance the suffix and remaining time prediction of event sequences. To capture the joint temporal dynamics of events, we harness the power of adversarial learning techniques to boost prediction performance.
We consider four real-life datasets and three baselines in our experiments. The results show improvements up to four times compared to the state of the art in suffix and remaining time prediction of event sequences, specifically in the realm of business process executions. We also show that the obtained improvements of adversarial training are superior compared to standard training under the same experimental setup.\footnote{The code of our tool can be obtained via \url{https://github.com/farbodtaymouri/MLMME}.}

\end{abstract}


\section{Introduction}
\label{sec: introduction}


Timestamped event sequences (event sequences for short) contain ordered events which carry at least two attributes: the event's label and its timestamp. The event label may indicate the label of a corresponding activity being executed, e.g.\ the activity of a business process; the event timestamp indicates when the event has been recorded, e.g.\ capturing when the corresponding process activity has been completed. 

Two common prediction tasks for event sequences are event suffix and remaining time prediction. Given a prefix, i.e.\ a partly-complete event sequence, suffix prediction aims to predict the most likely continuation of that prefix, while remaining time prediction aims to predict the sequence's remaining time, or the time that it will take to complete the most likely suffix. 

These two prediction tasks have various applications, for example in economics, digital health, business process management and IT infrastructure monitoring. For example, in the latter area, suffix and remaining time prediction help determine how well a given process execution (a so-called process case, e.g.\ an order, a purchase request or a claim) will be performing with respect to its performance measures and performance objectives \cite{DumasRMR18}. 
In IT infrastructure monitoring, a cloud system that monitors computation tasks and allocates resources to them can establish their future needs several steps ahead and prepare the required resources earlier than needed. Therefore, the overall performance of the system and quality of services can be improved.

\begin{figure}[h]
	\centering
	\includegraphics[width=.8\linewidth]{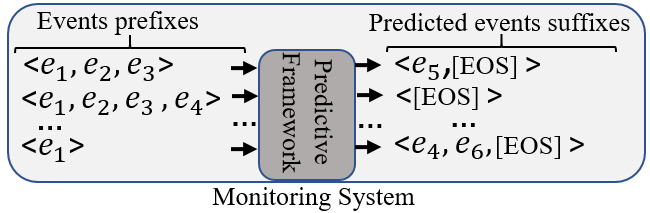}
	\caption{Suffix and remaining time prediction in a cloud system; Each $e_i  = (a_i,t_i)$.}
	\label{fig:IS}
	\vspace{-3mm}
\end{figure}
In this paper, we are specifically interested in the following problems: given an ongoing sequence of events, called the events prefix, and an event log of completed sequences, 
we want to predict the most likely continuation for that events prefix, by determining the sequence of events labels 
called the suffix, and the corresponding remaining time (total duration time of the events suffix) until the case finishes.  Figure \ref{fig:IS} depicts the problem we address in this paper, where an event $e_i=(a_i,t_i)$ has two components, i.e., the label and the timestamp. The sequence's end is denoted by [EOS].

Recently, suffix and remaining time prediction of timestamped event sequences have received great attention in the process mining field, where said sequences represent business process executions. Various approaches have explored these two prediction tasks, e.g.\ \cite{EVERMANN2017129,Tax17,Lin2019MMPredAD,Camargo2019LearningAL}, chiefly using machine learning models based on Recurrent Neural Networks (RNNs) with Long-Short-Term Memory (LSTM). By considering event logs recording completed process executions (also known as cases), train a machine learning model to predict the next event given an event prefix. Thus, the suffix of events is predicted by guessing the next event iteratively. These approaches are prone to potentially large prediction errors due to the discrepancy between closed-loop training  (i.e., the next event is conditioned on the ground truth of previous events) and open-loop inference (i.e., the next event is conditioned on previous predicted events). 
In addition, these approaches are limited in predicting a single suffix for an input prefix rather than providing a set of candidate suffixes. Such a set of predictions can be useful for further analysis, making recommendations, and monitoring possible risks in the future \cite{ProcessForecastingquteprints131983}.




The sequence to sequence learning task has been addressed in other fields.
 For example, Nallapati et al. \cite{Nallapati2016AbstractiveTS, Nallapati2017SummaRuNNerAR} use RNN-based architecture to learn a mapping between a document and its summarization. The works in \cite{SutskeverSEQtoSEQ, Vaswani2018Tensor2TensorFN, Bahdanau2015NeuralMT} use RNNs to propose various machine translation architectures that translates a text from a source language to its counterpart in a target language.  Such methods consider time-independent events. In particular, an event is a word or a character in a sentence. In contrast, in this paper,  an event is time-dependent, i.e., it has a timestamp that shows its duration time. Such timestamps are essential since they help to approximate occurrences of events and predict when a process execution finishes.



This paper proposes an encoder-decoder framework to learn a mapping from a set of events prefixes to a set of events suffixes in an end-to-end way where the events are time-dependent. It allows one for open-loop training which reduces suffix and remaining time prediction errors. However, open-loop training does not capture the joint temporal dynamics of events in ground truth suffixes. Motivated by Generative Adversarial Nets (GANs) \cite{GANNIPS2014_5423}, we harness the power of adversarial training to learn the joint probability distribution of events suffixes given events prefixes. As such, the approach helps to capture the joint temporal relationships among ground truth events' suffixes in order to improve the accuracy of suffix and remaining time prediction. 



In summary, the contribution of this paper is twofold:
\begin{itemize}
    \item  An encoder-decoder architecture for open-loop training to reduce suffix and remaining time prediction error;
    \item An adversarial training method to capture joint temporal dynamics of events across time in order to boost  prediction performance.
\end{itemize} 

We implemented our approach in an open-source library and used this to conduct a battery of experiments using four real-life datasets and three baselines from the realm of process mining.

The rest of this paper is organized as follows. The related work is provided in Sec. \ref{sec: related work}. The problem definition is presented in Sec. \ref{sec: basic concepts}. Section \ref{sec: the proposed framework} presents the approach while the evaluation is discussed in Sec. \ref{sec:experiment}. Finally, Sec. \ref{sec: conclusion} concludes the paper and sketches some ideas for future work.

\section{Related Work}
\label{sec: related work}
In this section we discuss previous approaches for suffix and remaining time prediction of event sequences, with a specific focus on works in process mining. Next, we discuss background work in the area of Generative Adversarial Networks for temporal data, which underpin our approach.

\subsection{Suffix and Remaining Time Prediction of Business Process Executions}
\noindent
In the context of process mining, LSTM architecture has been the main tool for various kinds of sequence prediction tasks. Evermann et al. \cite{EVERMANN2017129} use the LSTM architecture for the next event prediction of an ongoing case. It uses embedding techniques to represent categorical variables. Tax et al. \cite{Tax17} propose a similar architecture based on LSTMs using a one-hot vector encoding to represent categorical variables. Suffix prediction is made by next event's label predictions iteratively using the arg-max operator, i.e., the most probable activity, it outperforms \cite{EVERMANN2017129} in terms of next activity, suffix, and remaining time predictions. Similarly, Camargo et al. \cite{Camargo2019LearningAL} use embedding techniques similar to \cite{EVERMANN2017129}. This approach unlike the work in \cite{Tax17, EVERMANN2017129} uses other existing data attributes in process executions for the prediction tasks. Lin et al. \cite{Lin2019MMPredAD} propose a framework based on two LSTMs. This approach uses a modulator mechanism that utilizes all available information in input log, i.e., both event labels and performance attributes. The experiments show that this approach outperforms  \cite{EVERMANN2017129, Tax17, Pasquadibisceglie2019UsingCN}, and \cite{Camargo2019LearningAL} on some datasets.

Theis et al. \cite{Theis2019DecayRM} train a fully connected model that predicts the next event only. The experiments show an improvement over \cite{Tax17, EVERMANN2017129}. The works in \cite{Pasquadibisceglie2019UsingCN, DiMauroCNN} uses a Convolutional Neural Network (CNN) using image-like data structure for the next event's label prediction task in a running process execution. The experiments show an improvement over RNN-based architectures \cite{Tax17, EVERMANN2017129}. Taymouri et al. \cite{Taymouri2020PredictiveBP} propose a GAN architecture by invoking an LSTM for both the discriminator and the generator. The results showed that it outperforms previous techniques \cite{Tax17,EVERMANN2017129, Camargo2019LearningAL, Pasquadibisceglie2019UsingCN} for the next activity and timestamp prediction.

\subsection{Generative Adversarial Networks for Temporal Data} Adversarial   learning   has  received   considerable  attention   due to  its  capability  of  generating  the  synthetic  samples  that are  similar  to  the  real  one \cite{GANNIPS2014_5423}. Recently GANs have been shown to be effective in improving the temporal dynamics of autoregressive models for time-series data \cite{NIPS20198789GANtimesereis}.  Some works directly employ unsupervised objective by applying GANs framework to sequential data, mainly by instantiating recurrent networks for the roles of generator and discriminator. For example, Esteban et al. \cite{Esteban2017RealvaluedTimesereisGANs} propose Recurrent Conditional GAN (RCGAN) to produce realistic real-valued multi-dimensional time series for medical applications.
The work in \cite{Mogren16}  proposes (Continuous RNN-GAN) adversarial training using LSTM networks for generator and discriminator to model the whole joint probability of a real-values sequence that can be used for music production. On the other hand, the recent work \cite{NIPS20198789GANtimesereis}, combines the flexibility of the unsupervised paradigm with the  supervised  training for various continuous-valued time-series prediction tasks. Indeed, supervised and adversarial objectives encounter the network to adhere to the dynamics of the training data during training. Experiments show this approach outperforms baselines in  predictive ability of time-series.

While these methods' motivation is similar to ours in accounting for various temporal data prediction tasks, they consider continuous-valued time-series data. Indeed despite compelling results, little attention has been paid time-dependent events that are discrete-values.

\section{Problem Statement}
\label{sec: basic concepts}
In this section, we first define the preliminaries, and then we present the suffix and remaining time prediction formally.

In the suffix and remaining time prediction, the dataset, i.e., event log, is a set of sequences (process executions or traces) $L = \{\sigma^{(1)}, \sigma^{(2)},\dots, \sigma^{(l)} \}$, where $l$ is the size of dataset. The $i$-th process execution $\sigma^{(i)} = \langle e_1,e_2,\dots,e_n \rangle$ contains a sequence of $n$ events, i.e., $|\sigma^{(i)}| =n$. An event $e_j = (a_j,t_j)$ has two attributes where the former is the event's label, i.e., an activity, and the latter is the event's duration time, i.e., the required execution time. For a given process execution $\sigma^{(i)}$=$\langle (a_1,t_1),(a_2,t_2),\dots, (a_m,t_m) \rangle$, the prefix of events of length $k$ is defined by $\sigma^{(i)}_{\leq k}$=$\langle (a_1,t_1),(a_2,t_2),\dots, (a_k,t_k) \rangle$ and its corresponding suffix of events is $\sigma^{(i)}_{> k}$=$\langle (a_{k+1}, t_{k+1}), (a_{k+2}, t_{k+2}),\dots, (a_{m}, t_{m}) \rangle$.

\begin{Definition}(\textbf{Suffix prediction and remaining time prediction}) 
Suppose that there are pairs sample of sequences 
$\mathcal{S} = \{(\sigma_{\leq k}^{(i)}, \sigma_{> k}^{(i)}) \}_{i=1} ^{i=n}$, where $2 \leq k < |\sigma^{(i)}|$ is the prefix length and $n$ is the sample size.
Given a prefix of events sequence $\sigma_{\leq k}$=$\langle (a_1,t_1),(a_2,t_2),\dots, (a_k,t_k) \rangle$, 
the output prediction is the sequence of events $\widehat{\sigma}_{> k} = \langle (a_{k+1}, t_{k+1}), (a_{k+2}, t_{k+2}),\dots, [EOS] \rangle$, where $[EOS]$ is a special symbol added to the end of each process execution to mark the end of the sequence in prepossessing time. 
Suffix prediction is the sequence of activities in $\widehat{\sigma}_{> k}$, i.e., $\langle a_{k+1}, a_{k+2},\dots, [EOS] \rangle$. The remaining time prediction is the sum of the predicted duration time $t$ in $\widehat{\sigma}_{> k}$, i.e., $\sum t_i$, where $t_i \in \widehat{\sigma}_{> k}$.
\label{def:suffix and remaintime prediction}
\end{Definition}

\begin{figure}[h]
\vspace{-5mm}
	\centering
	\includegraphics[width=1\linewidth]{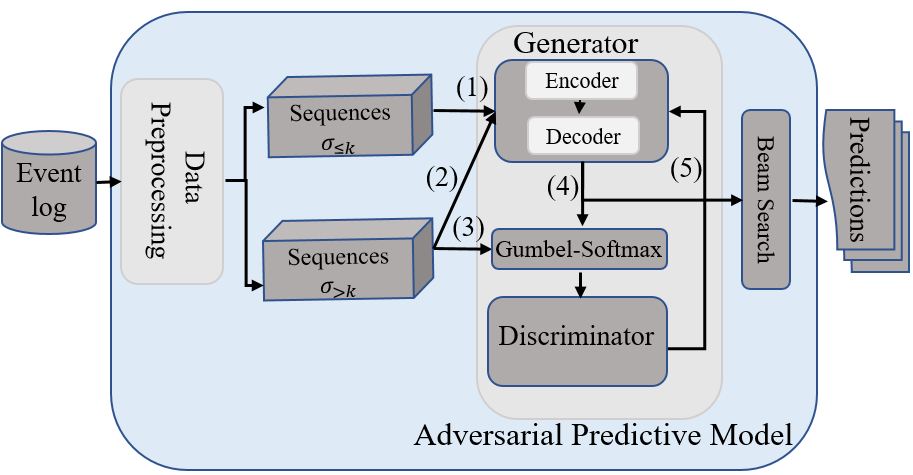}
	\vspace{-5mm}
	\caption{Framework for suffix and remaining time prediction}
	\label{fig:framework}
	\vspace{-5mm}
\end{figure}

\section{Proposed Framework}
\label{sec: the proposed framework}
In this section,  we  will  first give  an  overview of the proposed framework and then detail each component.


The proposed framework is shown in Fig. \ref{fig:framework}. Its objective is to find a mapping between event prefixes and event suffixes such that the predictions $\widehat{\sigma}^{(i)}_{> k}$ get close to the corresponding ground truths $\sigma^{(i)}_{> k}$, for suffix and remaining time prediction tasks.
To achieve this goal, the proposed approach exploits both the standard training using Maximum Likelihood Estimation (MLE), and the adversarial training inspired by GANs \cite{GANNIPS2014_5423}. It has three parts as follows:
\begin{itemize}
    \item \emph{Data prepossessing}: It prepares the input data in the form of events prefixes and suffixes for the suffix and remaining time prediction tasks. 
    \item \emph{Adversarial predictive model}: It is made of two neural networks, i.e., the generator and the discriminator, where the former provides predictions, i.e., $\widehat{\sigma}_{> k}$, and the latter evaluates the predictions using ground truth $\sigma_{> k}$. The generator receives feedback from the ground truths directly, i.e., supervised loss, and the discriminator's output, i.e., adversarial loss.
    \item \emph{Beam search}: To have more than one suffix and remaining time prediction for an events prefix, we employ beam search on top of the generator’s outputs to provide several candidates, i.e., suffix and remaining time predictions. The number of candidates is determined by the beam size $n$.
\end{itemize}

\subsection{Data Preprocessing}
In the proposed framework each event $e_i = (a_i, t_i)$ is shown by vector $\mathbf{e}^{(i)} = (\mathbf{a}^{(i)}, t_i)$, where $\mathbf{a}^{(i)}$ is the one-hot encoding of activity $a_i$. We denote the $j$-th entry of a vector by the corresponding subscript, e.g., $\mathbf{a}^{(i)}_j$. 
With such representation, an events prefix or an events suffix is shown by a sequence of vectors. For example, given $\sigma = \langle \mathbf{e}^{(1)},\mathbf{e}^{(2)},\mathbf{e}^{(3)},\mathbf{e}^{(4)} \rangle$, $\sigma_{\leq 2} = \langle \mathbf{e}^{(1)},\mathbf{e}^{(2)} \rangle$, $\sigma_{> 2} = \langle \mathbf{e}^{(3)},\mathbf{e}^{(4)} \rangle$, and $\sigma_{\leq 3} = \langle \mathbf{e}^{(1)},\mathbf{e}^{(2)},\mathbf{e}^{(3)} \rangle$, $\sigma_{> 3} = \langle \mathbf{e}^{(4)} \rangle$.

Since $\mathbf{e}^{(i)}$ is a vector containing representations for $e_i = (a_i, t_i)$, we define two functions $f_a()$ and $f_t()$ which extract the relevant $a_i$ and $t_i$ in $\mathbf{e}^{(i)}$. These functions can be applied to sequence of events as well, e.g., $f_a(\sigma_{\leq 2}) = \langle a_{1},a_{2} \rangle$ and $f_t(\sigma_{> 2}) = \langle t_3,t_4 \rangle$.

\subsection{Adversarial Predictive Model}
In this part, we first provide the detail of the proposed encoder-decoder architecture and its supervised training. Next, we explain the proposed adversarial training.

\begin{figure}[h]
	\centering
	\includegraphics[width=.9\linewidth]{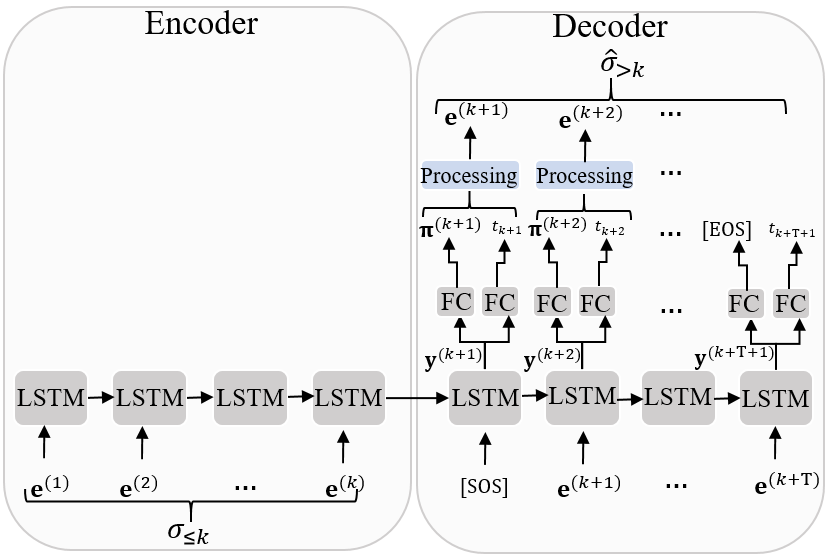}
	\vspace{-3mm}
	\caption{Encoder-decoder architecture}
	\label{fig:encoder-decoder}
	\vspace{-1mm}
\end{figure}

\noindent \textbf{Encoder-decoder framework}: Figure \ref{fig:encoder-decoder} shows the proposed encoder-decoder architecture in this paper. In this framework, Long Short-Term Memory (LSTM) \cite{LSTM1997} is used for both the encoder and the decoder. Moreover, the decoder contains two additional fully connected (FC) layers that share LSTM cells for predicting an activity and its duration time. In that framework, [SOS] is the symbol for starting the prediction.
In detail,
the encoder-decoder architecture in Fig. \ref{fig:encoder-decoder}, provides $\widehat{\sigma}_{> k}$ as the prediction for the ground truth $\sigma_{> k}$. The encoder maps the events prefix into a vector and feeds it to the decoder. The decoder iteratively predicts next events.
Formally, suppose that $\mathbf{y}^{(k+1)}$ is the output of LSTM for step $k+1$, then:
\begin{equation}
\footnotesize{\begin{aligned}
    & \boldsymbol{\pi}^{(k+1)} = Softmax(\mathbf{W}_a\mathbf{y}^{(k+1)} + \mathbf{b}_a) \\
    & {t}_{k+1} = ReLU(\mathbf{W}_t\mathbf{y}^{(k+1)} + \mathbf{b}_t) 
\end{aligned}}
\label{eq:softmax}
\end{equation}

\noindent where the Softmax function \cite{Goodfellow-et-al-2016} provides $\boldsymbol{\pi}^{(k+1)}$, i.e., the predicted probability vector for activities including [EOS], and $t_{k+1}$ is the predicted duration time. $\mathbf{W}_a, \mathbf{W}_t, \mathbf{b}_a$, and $\mathbf{b}_t$ are the parameters of FCs. Note that the Rectified Linear Unit (ReLU) in Eq. \ref{eq:softmax} provides a non-negative output. 

The Processing block in Fig. \ref{fig:encoder-decoder} creates one-hot encoding for an activity with the highest probability. It then concatenates that vector with the corresponding predicted duration time $t_{k+1}$, which results in $\mathbf{e}^{(k+1)}$. After that, it is fed as input for the next step prediction, and thus open-loop training can be achieved.



If the ground truth activity at step $k+1$ is $\mathbf{a}^{(k+1)}_i$, i.e., the $i$-th entry, and the network's prediction for such activity is $\boldsymbol{\pi}_i^{(k+1)}$ then the corresponding error, also known as the cross-entropy, is $L^{(k+1)} = - \mathbf{a}^{(k+1)}_i log \boldsymbol{\pi}_i^{(k+1)}$. Thus, the suffix prediction task's loss function is the sum of such errors over all time steps until reaching [EOS]:

\begin{equation}
    \footnotesize{L_{activity} = \sum_{j} L^{(j)}}
    \label{eq: activity loss}
\end{equation}

\noindent In addition, the error of remaining time prediction is the squared difference between the ground truth and the predicted remaining times. Formally:
\begin{equation}
    \footnotesize{L_{time} = (\sum f_t(\widehat{\sigma}_{> k}) - \sum f_t(\sigma_{> k}))^2}
    \label{eq: timeloss}
\end{equation}
Finally, one can train the proposed encoder-decoder by minimizing the weighted sum of Eq. \ref{eq: activity loss} and \ref{eq: timeloss}, i.e., the supervised loss, for the training set:
\begin{equation}
    L_{supervised} = w_aL_{activity} + w_tL_{time}
    \label{eq: supervised loss}
\end{equation}

\noindent where $w_a $, and $w_t$ are tunable positive weights which, for the sake of simplicity, we consider $w_a = w_t = 1$.

\noindent \textbf{Adversarial Training}:
The adversarial training in this paper is motivated by GANs \cite{GANNIPS2014_5423}. 
In this training procedure, as shown in Fig. \ref{fig:framework}, we call the encoder-decoder architecture as the generator and denote it by $G(;\mathbf{\theta}_g)$. The discriminator, denoted by $D(;\mathbf{\theta}_d)$, is composed of LSTM followed by a fully connected layer. The trainable parameters of the generator and the discriminator are denoted by $\theta_g$ and $\theta_d$, respectively. In detail, given a prefix of events, i.e., $\sigma_{\leq k}$, the generator's output is a sequence of events, i.e., $G(\sigma_{\leq k})= \hat{\sigma}_{> k}$.

The adversarial training works as a minmax game between the generator and the discriminator. It starts
by proposing fake and real instances. A real instance is a suffix of events in the training set, i.e., $\sigma_{> k}$, and a fake instance is formed from the generator's output, i.e., $\hat{\sigma}_{> k}$. The training runs as a game between two players, where the generator's goal is to maximize
the quality of predictions, i.e., accurate suffix and remaining time prediction, to fool the discriminator. The discriminator's goal is to minimize its error by evaluating the quality of generator's predictions, see flow 4. In particular, the discriminator assigns high probability values to real instances and low probability values to fake instances. It is an adversarial game since the generator and the discriminator compete with each other, i.e., learning from the opponent's feedback, see flows (4), (5) in Fig. \ref{fig:framework}, thus maximizing one objective function minimizes the other one and vice versa.

The discriminator can send feedback to the generator, i.e., updating the generator's parameters via backpropagation, if its inputs are differentiable \cite{GANNIPS2014_5423}; however in our work both $\sigma_{> k}$, i.e., the ground truth, and the generator's output, i.e., $\hat{\sigma}_{> k}$,  contain categorical items, i.e., activity, that have zero gradients with respect to $\theta_g$ and $\theta_d$, respectively. Thus, we get a continuous approximation to such categorical items using Gumbel-softmax distribution \cite{gumbel1954statistical,NIPS2014_5449Gumble}. In particular, if $\boldsymbol{\pi}^{(k+1)}$ shows the vector of probability prediction of activities for step $k+1$, see Eq. \ref{eq:softmax}, the continuous approximation is the vector $\boldsymbol{\alpha}^{(k+1)}$, with the $i$-th entry as follow:

\begin{equation}
\label{eq: gumble softmax}
   \footnotesize{ \boldsymbol{\alpha}_i^{(k+1)} = \frac{exp((log (\boldsymbol{\pi}_i^{(k+1)}) + g_i)/\tau)}{\sum_{j=1}^m exp((log (\boldsymbol{\pi}_j^{(k+1)}) + g_j)/\tau)} } 
\end{equation}


\noindent where  $g_i$ are i.i.d samples from Gumbel(0,1), $m$ is the number of activities including [EOS], and $\tau$ is a parameter called \emph{temperature}. When $\tau \rightarrow 0$, the entries of $\boldsymbol{\alpha}$, look like the one-hot encoding representation, while $\tau \rightarrow \infty$ the corresponding entries constitute a vector with uniform probability values.


The Gumbel-softmax part in Fig. \ref{fig:framework} creates event vectors $\mathbf{e}^{(i)} = (\boldsymbol{\alpha}^{(i)}, t_i)$ for the generator's output using Eq. \ref{eq: gumble softmax}, and for every one-hot vector $\mathbf{a}^{(i)}$ in $\sigma_{>k}$ it forms a continuous approximation vector by placing a high probability, e.g., 0.9, on the correct activity, and $(1-0.9)/(m-1)$ on the remaining activities, where $m$ is the number of activities.



In the minmax game, we want the $G$'s output, i.e., $\hat{\sigma}_{> k}$, to be as close as possible to ground truth $\sigma_{> k}$, such that, $D$ gets confused in discriminating the mentioned suffixes of events. Formally, for a pair $(\sigma_{\leq k}, \sigma_{>k}))$ and the prediction $G(\sigma_{\leq k})= \hat{\sigma}_{> k}$, we consider the following adversarial loss functions for the discriminator and the generator:

\begin{figure}
	\centering
	\includegraphics[width=1\linewidth]{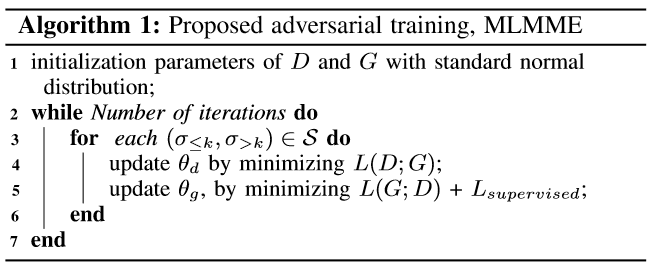}
	\vspace{-11mm}
\end{figure}

\begin{flalign}
\begin{split}
\label{eq: gan equation}
   L(D;G) = -log(D(\sigma_{>k})) - log(1-D(\hat{\sigma}_{> k})) \\
    L(G;D) = -\left [ log(D(\hat{\sigma}_{> k}))  - log(1-D(\hat{\sigma}_{> k})) \right ]
\end{split}
\end{flalign}


\noindent  Equation \ref{eq: gan equation} iterates two steps:  first, it updates discriminator $D$ by lowering  $L(D;G)$, keeping $G$ fixed, then it updates $G$ by lowering $L(G;D)$ keeping $D$ fixed. It can be shown that the optimization in Eq. \ref{eq: gan equation} amounts to minimizing the distance between two probability distributions that generate the ground truth event suffixes and the predicted event suffixes, respectively \cite{Sonderby2016aGAN}. Thus, one captures the joint temporal dynamics of events in ground truth events suffixes.


The proposed adversarial training which we call Maximum Likelihood Min-Max Estimation (MLMME) is shown in Alg. 1. For each pair of prefix and suffix $(\sigma_{\leq k}, \sigma_{>k})$ in the training set $\mathcal{S}$, we update the parameters of the discriminator and the generator according to Eq. \ref{eq: gan equation}. Also, the generator's parameters are further updated using Eq. \ref{eq: supervised loss}, i.e., supervised loss. Note that, unlike original GAN \cite{GANNIPS2014_5423}, which needs Nash equilibrium to stop training, we consider a specific number of iterations in Alg. 1. Indeed, the objective of Eq. \ref{eq: gan equation} is to help the supervised training to boost the generator's performance in learning joint temporal dynamics of events across time to advance suffix and remaining time prediction.
After training, we disconnect the discriminator and use the generator for the prediction tasks.

\subsection{Beam Search}
\label{subsec:beam search} After training and during inference, the suffix and remaining time prediction proceeds one step at a time.
At each step, we predict one output, i.e., the activity $a_i$, and its duration time $t_i$, see Fig. \ref{fig:encoder-decoder}. We first compute a probability
distribution over all activities using Eq. \ref{eq:softmax}. We then pick the most likely activity and move to the next prediction step. This process is 1-best greedy search and makes us vulnerable to the so-called Garden-Path problem \cite{GardenPath1988}. In that
case, the best predicted suffix consists initially of less probable activities, which
are redeemed by subsequent activities in the output sequence.

To alleviate this issue, when predicting the first activity of the decoder output, we keep a beam of the top $n$, called beam size, most likely activity choices. They are scored by
their probability. Then, we use each of these activities in the beam in the conditioning context for the next activity. Due to this conditioning, we
make different predictions for each. We now multiply the score
for the partial prediction, and the probabilities from its activity predictions. We select the
highest scoring activity pairs for the next time step. This process continues. At each time step, we accumulate the output activity probabilities, giving us scores for each partial prediction. A prediction is complete when the end of sequence token, i.e., [EOS], is produced. At this point, we remove the completed prediction, i.e., $\hat{\sigma}_{> k}$, from the beam and reduce beam size by 1. Search terminates, when no partial predictions are left
in the beam.

A Beam search with size $n$, returns the $n$ most probable suffix predictions for an input event prefix among those explored in the search process. The time and space complexities of Beam search with beam size $n$ are $\mathcal{O}(nb)$ and $\mathcal{O}(n)$, respectively, where $b$ is the branching factor or the number of activities in our work.

\section{Experiments}
\label{sec:experiment}
In this section, we will first introduce the datasets, baselines, and experimental settings. Next, we compare our results to those of baselines for suffix and remaining time prediction. Note that, since all the baselines use 1-best greedy search for suffix and remaining time prediction, we adopt the same strategy for the comparison.
First, we show that the proposed method outperforms baselines.
Next, we investigate the strength of the proposed adversarial training. In particular, using various beam sizes, we show the efficiency of the proposed adversarial training method, i.e., MLMME, against the standard training method, i.e., MLE, for the suffix and remaining time prediction tasks.

\subsection{Datasets}
We evaluate the performance of the proposed technique for suffix and remaining time prediction on four real-life datasets that are publicly available. No prepossessing are applied to datasets. All detests were used in baselines \cite{Tax17, Taymouri2020PredictiveBP, Lin2019MMPredAD}. Table \ref{table:dataset} shows some characteristics of the datasets. For each dataset, \#Sequence shows the number of sequences, i.e., process executions, \#Events is the number of events, \#Activity represents the number of unique activities, Sequence-Length provides the length of shortest and longest sequences, and the last column shows the average and maximum cycle time, i.e., the time from starting a process execution until it ends.  In details:

\begin{itemize}
    \item Helpdesk:\footnote{\url{https://data.4tu.nl/articles/dataset/Dataset_belonging_to_the_help_desk_log_of_an_Italian_Company/12675977}} It contains sequences from a ticketing management process of the help desk of an Italian software company. 
    All traces start with the insertion of a new ticket into the ticketing management system. Each case ends when the issue is resolved and the ticket is closed in management system.
    \item BPI12\footnote{\url{https://data.4tu.nl/articles/dataset/BPI_Challenge_2012/12689204}}: It contains sequences of a loan application process at a  Dutch financial institute  through  an  on-line  system  from  2011/10/01  to  2012/03/14. This process includes three sub-processes from which one of them is denoted as $W$ and used already in \cite{Tax17, Taymouri2020PredictiveBP}. As such, we extract it from this dataset, i.e., BPI12(W).
    \item BPI17:\footnote{\url{https://data.4tu.nl/articles/dataset/BPI_Challenge_2017/12696884}} It contains sequences of a loan application process at the same Dutch financial institute in BPI12 but for 2016 and their subsequent handling up to February 2nd, 2017.
\end{itemize}

\begin{table*}[h]
    \centering
\footnotesize{\begin{tabular}{|c |c |c |c |c |c |}
  \hline
  \textbf{Dataset}  & \textbf{\#Sequence} & \textbf{\#Events} & \textbf{\#Activity} & \textbf{Sequence Length} & \textbf{Avg. - Max Cycle Time (day)} \\ 
  \hline
  Helpdesk & 3,804     & 13,087 & 9  & 1-14 & 8.79 - 55.9 \\
  BPI12 & 13,087     & 262,200 & 23  & 3-96 & 8.6 - 91.4\\
  BPI12(W) & 9,658     & 72,413 & 6  & 1-74  & 11.4 - 91 \\
  BPI17 & 31,509     & 1,202,267 & 24  & 7-54 & 21.8 - 169.1\\
  \hline
\end{tabular} }
\caption{Descriptive statistics of the datasets}
\label{table:dataset}
\vspace{-5mm}
\end{table*}

For  each  dataset,  we  consider temporal splitting into  three  sets  by 7:1:2  ratio.   We  use  the  first  70\% of  the  sequences  for training, the middle 10\% of  sequences  for validation  and  the  remaining  20\% sequences for evaluating the performance of the model after training.

\subsection{Experimental Setup} The proposed approach is implemented in Python 3.7, Pytorch 1.2, and CUDA 10.1 on a Linux server using two NVIDIA P100 GPUs with 64 GB RAM.
For the encoder, the decoder, and the discriminator we use a five layer LSTM with 200 neurons in each layer. In addition, the discriminator is equipped with a fully connected layer. In detail:
\begin{itemize}
    \item We consider 500 iterations using the proposed adversarial training, i.e., MLMME. To speed-up the training convergence, we used a probabilistic teacher forcing ratio of 0.1, which accordingly allows the decoder to use the ground truth from a prior time step as input \cite{Goodfellow-et-al-2016};
    
    \item For each dataset, i.e., event log, a training instance is a pair of prefix and suffix events $(\sigma_{\leq k}, \sigma_{>k}))$, where the prefix length, i.e., $k$, is equal to or greater than 2; 
    \item We apply early stopping if we observe no more improvement on the validation set for 30 iterations;
    \item We use RMSprop as an optimization algorithm for the proposed framework with learning rate ${5e-5}$. To avoid gradient explosion, we clip the gradient norm of each layer to 1; and 
    \item We exponentially anneal the temperature $\tau$ of the Gumbel-Softmax distribution from 0.9 to 0 in Eq. \ref{eq: gumble softmax} to stabilize the training.
\end{itemize}

\noindent 

\noindent \textbf{Baselines:} We consider three baselines according to their competitive performances, i.e., \cite{Tax17, Lin2019MMPredAD, Taymouri2020PredictiveBP}. Taymouri et al. \cite{Taymouri2020PredictiveBP} predicts the next event only. However, we developed a script for event suffix prediction by  feeding the outputs of events  back  as  the  input  of  the  predictive  model repeatedly to obtain suffix and remaining time prediction. The recent work by Lin et al. \cite{Lin2019MMPredAD} predicts all the categorical attributes (and not only the activity label) of the future events, but not their
associated timestamps In addition, the toolbox of this technique is not publicly available so we compare with the reported results in the respective paper. For \cite{Tax17, Taymouri2020PredictiveBP}, we used the best parameter settings for training, as discussed in the respective papers.

\noindent \textbf{Evaluation measures:} To  evaluate  the  performance  of  suffix  prediction, we  use  Damerau-Levenstein  distance  (DL) \cite{DamerauDistance}. This metric measures the quality of the predicted suffix by adding swapping operation to the set of operations used by Levenstein distance. For example, for two sequences $\langle a_1,a_2,a_3 \rangle$ and $\langle a_1,a_3,a_2 \rangle$, it assigns a cost of 1.0 for swapping $a_2$ and $a_3$. Given two activity sequences $s_1 = f_a(\sigma_1)$ and $s_2=f_a(\sigma_2)$ we consider the following similarity:
\begin{equation}
\label{eq: dl metric}
    \footnotesize{SDL(s_1,s_2) = 1-\frac{DL(s_1,s_2)}{Max(len(s_1),len(s_2))}}
\end{equation}

\noindent Where $len(s)$ is the length of $s$, i.e., the number of elements in $s$. The metric in Eq. \ref{eq: dl metric} is also employed by baselines \cite{Tax17, Lin2019MMPredAD}. $SDL \in [0,1]$, and it is 1.0 when two sequences are the same and 0.0 when two sequences contain completely different elements. Also, we compute the absolute error (AE) between the ground truth remaining time and the predicted remaining time for each predicted and ground truth sequences. Next, we averaging these numbers for test instances, and report Mean Absolute Error (MAE).

\subsection{Experimental Results}
\label{sec:results}

\begin{table*}[h]
	\centering 
	\footnotesize{ \begin{tabular}{|p{2.6cm}| p{1.2cm}| p{1.3cm} | p{1.0cm} | p{1.0cm} | p{1.2cm}| p{1.3cm}|p{1.0cm}|p{1.0cm}|}
		\hline
    \multicolumn{1}{|c|}{} & \multicolumn{4}{|c|} {\textbf{Average SDL}} & \multicolumn{4}{|c|}{\textbf{ MAE (day)}}\\
		\hline \hline
		\textbf{Approach} & Helpdesk  & BPI12(W) & BPI12 & BPI17 & Helpdesk  & BPI12(W) & BPI12 & BPI17\\ [1ex]
		\hline
		Ours  & \textbf{0.8411}  &  \textbf{0.2662}    & \textbf{0.3326}  & \textbf{0.3361}  & \textbf{6.21}  & \textbf{12.12} & \textbf{13.62}  & \textbf{13.95} \\
		Taymouri et al. \cite{Taymouri2020PredictiveBP} &0.8089  &  0.2125  & 0.2266  & 0.2881  & 6.30  & 34.56 & 169.23  &  80.81 \\
		Tax et al. \cite{Tax17} &0.7670  &  0.0632  & 0.1652  & 0.3152  & 6.32  & 50.11 &  380.10  & 170.02 \\
		\hline
	\end{tabular} }
		\caption{Average  SDL for the suffix prediction (the larger, the better), and MAE for predicting remaining time.}
	\label{table:accuracy and MAE}
\end{table*}

\begin{table*}[h]
	\centering 
	\footnotesize{ \begin{tabular}{|p{2.6cm}| p{1.2cm}| p{1.3cm} | p{1.2cm} | p{1.0cm} | p{1.2cm}| p{1.3cm}|p{1.0cm}|p{1.0cm}|}
		\hline
    \multicolumn{1}{|c|}{} & \multicolumn{4}{|c|} {\textbf{P-value for the SDL comparison}} & \multicolumn{4}{|c|}{\textbf{ P-value for A.E. comparison}}\\
		\hline \hline
		\textbf{Approach} & Helpdesk  & BPI12(W) & BPI12 & BPI17 & Helpdesk  & BPI12(W) & BPI12 & BPI17\\ [1ex]
		\hline
		Taymouri et al. \cite{Taymouri2020PredictiveBP} &0.61e-9  &  0.25e-6  & 0.14e-12  & 0.11e-6  & 0.0351  & 0.0 & 0.0  &  0.0 \\
		Tax et al. \cite{Tax17} &1.2e-17  &  0.0  & 0.0  & 0.71e-4  & 0.0203  & 0.0 &  0.0  & 0.0 \\
		\hline
	\end{tabular} }
		\caption{P-values of paired t-tests for comparing our approach with baselines.}
	\label{table: p-value for accuracy}
\end{table*}

\begin{table}[h]
	\centering 
	\footnotesize{ \begin{tabular}{|p{2.3cm}| p{1.2cm}| p{1.3cm} | p{1.0cm} | p{1.0cm} | p{1.2cm}| p{1.3cm}|p{1.0cm}|p{1.0cm}|}
		\hline
    \multicolumn{1}{|c|}{} & \multicolumn{3}{|c|} {\textbf{Average SDL}}\\
		\hline \hline
		\textbf{Approach} & Helpdesk & BPI12 & BPI17 \\ [1ex]
		\hline
		Ours  & \textbf{0.8852}    & \textbf{0.4107}  & \textbf{0.3668} \\
		Lin et al. \cite{Lin2019MMPredAD} &0.8740   & 0.2810  & 0.3010 \\
		\hline
	\end{tabular} }
		\caption{Average SDL for the suffix prediction using experimental settings in \cite{Lin2019MMPredAD}.}
	\label{table:accuracy and MAE with Lin}
		\vspace{-7mm}
\end{table}

We first discuss the results on the two prediction tasks in terms of accuracy. Next, we study the effects of adversarial training on the results. Finally, we discuss the time complexity of the approach.

\subsubsection{Suffix and Remaining Time Prediction}
\label{subsec:suffix and remaining time prediction}
Table \ref{table:accuracy and MAE} shows the performance of the proposed approach against two baselines \cite{Taymouri2020PredictiveBP, Tax17} for 4 real-life logs. The first part of this table shows the predicted suffixes' quality according to the SDL metric, i.e., Eq. \ref{eq: dl metric}. One sees that the proposed method in this paper outperforms baselines and provides more accurate suffix predictions. The results witness the proposed approach can learn the complex dynamics of event across time much better than baselines. In particular, we got 4 percentage points improvement for the Helpesk dataset; 5 percentage points improvement for BPI12(W); and 11 and 4 percentage points improvement for BPI12 and BPI17, respectively. Similarly, Table \ref{table:accuracy and MAE with Lin} shows our approach outperforms Lin et al. \cite{Lin2019MMPredAD}. The experimental settings of this baseline is different than baselines \cite{Tax17,Taymouri2020PredictiveBP} in which prefixes containing at least 5 events (at least 4 events for Helpdesk) are considered for suffix predictions and evaluation. Thus, the average SDL metric increases since short prefixes could incur inaccurate suffix predictions. 

The second part of Table \ref{table:accuracy and MAE} represents the performance of our technique for remaining time predictions according to MAE values versus baselines \cite{Taymouri2020PredictiveBP, Tax17}. These results show the improvements are several times more accurate than baselines, and they become more recognizable for detests containing longer sequences. In particular, for BPI12(W) we got at least 250\% and 400\% improvement compared to Taymouri et al. \cite{Taymouri2020PredictiveBP} and Tax et al. \cite{Tax17}, respectively. One sees the approach in this paper by learning the complex dynamics of events across time can predict the remaining time of process executions more accurately at least several weeks for large datasets, e.g., see MAE values for BPI12.

\textbf{Statistical test}: To show that the improvements in suffix and remaining time prediction are obtained systematically, we apply paired t-tests.
In detail, we pair the SDL values of our approach with the corresponding SDL values provided by each baseline \cite{Taymouri2020PredictiveBP, Tax17}. By the same token, we pair the remaining time absolute errors provided by our technique with the corresponding values presented by each baselines. Moreover, we consider an upper-tailed t-test  
for paired SDL values where the null hypothesis states the average of differences is zero, and the alternative hypothesis states the average of differences is larger than zero, i.e., our approach is more accurate by providing larger SDL values. Likewise, we use another a lower-tailed test for paired remaining time absolute errors where the null hypothesis states the average remaining time absolute errors is zero, whereas  the alternative hypothesis states the average differences is less than zero, i.e., our approach provides smaller absolute errors. Table \ref{table: p-value for accuracy} shows the results of statistical tests between the approach in this paper and each baseline, i.e., \cite{Taymouri2020PredictiveBP, Tax17}, for datasets used in this paper. One sees that all the reported p-values are less than 0.05, meaning that the obtained improvements for suffix and remaining time predictions by our technique are statistically significant compared to baselines. We were unable to conduct such a statistical analysis for Lin et al. \cite{ Lin2019MMPredAD} due to the toolbox's unavailability.

\begin{figure*}[h]
\centering
	\includegraphics[width=.85\linewidth]{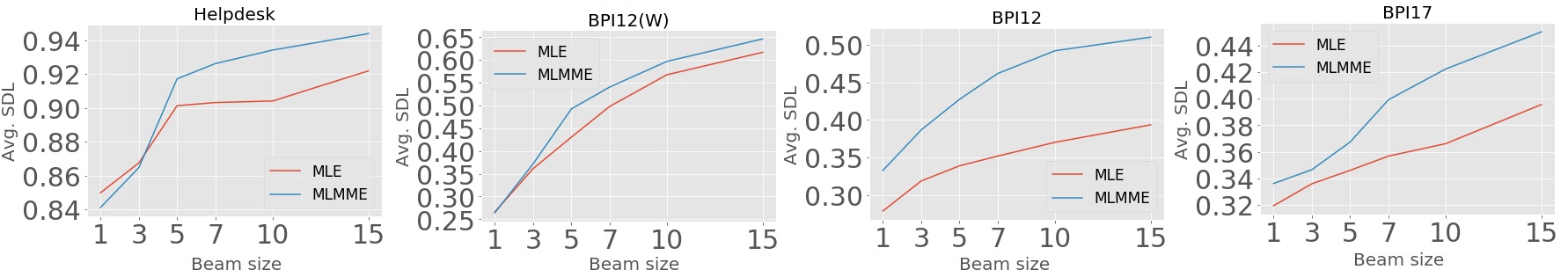}
	\vspace{-4mm}
	\caption{Average SDL values for various beam sizes between the conventional training, i.e., MLE, and the adversarial training, i.e., MLMME.}
	\label{fig:SDL comparison MLE MLMME}
\end{figure*}

\begin{figure*}[h]
\centering
	\includegraphics[width=.85\linewidth]{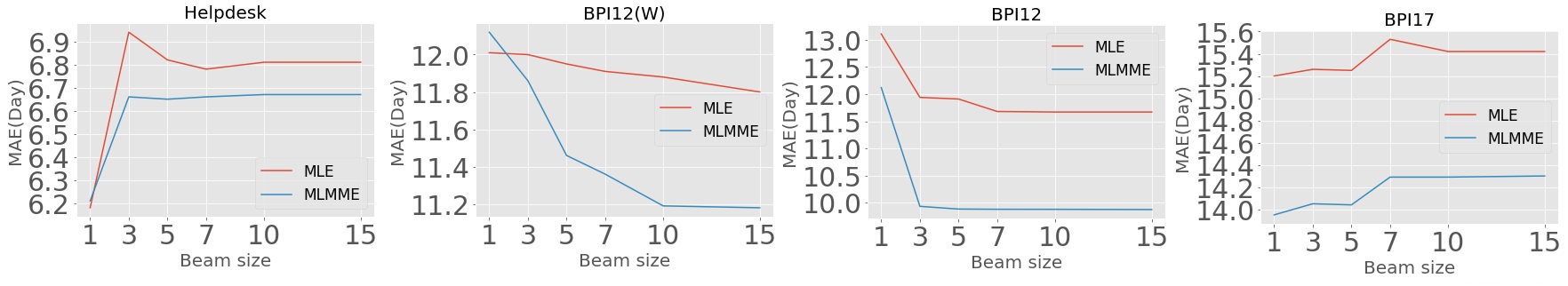}
	\vspace{-4mm}
	\caption{MAE values for various beam sizes between the conventional training, i.e., MLE, and the adversarial training, i.e., MLMME.}
	\label{fig:MAE MLE MLMME}
	\vspace{-1mm}
\end{figure*}

\begin{table*}[h!]
	\centering 
	\footnotesize{ \begin{tabular}{|p{2.2cm}| p{1.3cm}| p{1.5cm} | p{1.3cm} | p{1.5cm} | p{1.2cm}| p{1.3cm}|p{1.0cm}|p{1.0cm}|}
		\hline
    \multicolumn{1}{|c|}{} & \multicolumn{4}{|c|} {\textbf{P-value for the SDL comparison}} & \multicolumn{4}{|c|}{\textbf{ P-value for A.E. comparison}}\\
		\hline \hline
		\textbf{Beam Size} & Helpdesk  & BPI12(W) & BPI12 & BPI17 & Helpdesk  & BPI12(W) & BPI12 & BPI17\\ [1ex]
		\hline
		n=1 & 0.1132  &  0.2304  & \textbf{0.14e-12}  &\textbf{1.32e-10}  & 0.2601  & 0.0910 & \textbf{0.0}  &  \textbf{0.0} \\
		n=3 & 0.2302  &  \textbf{2.64e-6}  & \textbf{0.0}  & \textbf{8.77e-131}  & \textbf{0.0230}  & \textbf{2.07e-6} &  \textbf{0.0}  & \textbf{0.0} \\
		n=5 & \textbf{6.36e-7}  &  \textbf{3.31e-114}  & \textbf{0.0}  & \textbf{0.0}  & \textbf{0.0345}  & \textbf{0.0} &  \textbf{0.0}  & \textbf{0.0} \\
		n=7 & \textbf{1.42e-15} &  \textbf{2.71e-73}  & \textbf{0.0}  & \textbf{0.0} & \textbf{0.0381}  & \textbf{0.0} & \textbf{0.0}  & \textbf{0.0} \\
		n=10 & \textbf{1.17e-24}  &  \textbf{1.50e-48}  & \textbf{0.0}  & \textbf{0.0}  & \textbf{0.0225}  & \textbf{0.0} &  \textbf{0.0}  & \textbf{0.0} \\
		n=15 & \textbf{6.85e-17}  &  \textbf{4.15e-52}  & \textbf{0.0}  & \textbf{0.0} & \textbf{0.0265}  & \textbf{0.0} &  \textbf{0.0}  & \textbf{0.0} \\
		\hline
	\end{tabular} }
		\caption{P-values of paired t-test for comparing the performance of standard training, i.e., MLE, versus the adversarial training, i.e., MLMME.}
	\label{table: p-value for mlmme and mle}
		\vspace{-4mm}
\end{table*}

\subsubsection{Effects of Adversarial Training}
This section details the effects of the proposed adversarial training, i.e., MLMME, for suffix and remaining time prediction. To show that considering both the  adversarial loss and supervised loss in Eq. \ref{eq: supervised loss} and Eq. \ref{eq: gan equation}, respectively,  better capture the events temporal dynamics for suffix and remaining time prediction, we conduct the following experiment. Using the same experimental setup as outlined in the previous section we train another encoder-decoder network using only the MLE method, i.e., considering supervised loss in Eq. \ref{eq: supervised loss}. Next, we compare its performance against the model that is trained using MLMME to figure out which one provides more accurate suffix and remaining time prediction. In particular, to better shed light on each model's performance, we consider the suffix and remaining time prediction tasks using a beam search with various beam size $n \in \{1,3,5,7,10,15\}$.  Given a trained model, the beam search with beam size $n$ retrieves the $n$ best suffixes for each prefix, i.e., the $n$ most probable suffixes that the model can predict.
Therefore,  one can measure the largest SDL values and the corresponding remaining time absolute errors among the beam size candidates for a given prefix. Finally, if a model provides the larger average SDL value for the whole event prefixes, then it has captured the temporal dynamics of events across time more accurately.

Figure \ref{fig:SDL comparison MLE MLMME} shows the average SDL values of suffix prediction for various beam sizes and different datasets. One sees that the trained model using the MLMME can better learn the temporal dynamics of events across time. Therefore, it provides more accurate suffix predictions for a given prefix. Besides, by increasing the beam size it presents more accurate candidates for suffix predictions that can be used for recommendations or further analysis. Also, from Fig. \ref{fig:SDL comparison MLE MLMME}, the proposed adversarial training method excels for datasets containing long process executions, i.e., sequences, and having a large number of activities, e.g., see the SDL curves for datasets BPI12, BPI17. On the other hand, for datasets containing small number of activities, e.g., see Helpdesk and BPI12(W), the MLE training method shows a competitive performance for small beam sizes.

Figure \ref{fig:MAE MLE MLMME} represents the MAE values for the remaining time prediction for various beam sizes. One sees that the model with adversarial training, i.e., MLMME, provides smaller MAE values compared to the standard training, i.e., MLE. The effects of adversarial training become more effective for datasets containing long sequences and a large number of activities, e.g., BPI12, BPI17. On the other hand, the MLE method provides smaller MAE values for beam size equals to 1 for two datasets, i.e., BPI12(W) and Helpdesk. However, we will show that these differences are not statistically significant. 

It must be noted that for a given prefix and beam size $n$, the predicted suffix with the largest SDL value among the $n$ candidates does not necessarily have the lowest remaining time error prediction. This is because the beam search objective is to find more accurate suffix predictions by considering the activity probabilities among the best $n$ predicted suffixes, and it does not take into account the corresponding duration times associated with each activity. For example, in Fig. \ref{fig:MAE MLE MLMME} and \ref{fig:SDL comparison MLE MLMME} for two detests Helpdesk and BPI17, the average SDL values are decreased for larger beam sizes. However, MAE values are increased.

\textbf{Statistical test}: This section examines whether the superiority of adversarial training in suffix and remaining time prediction is statistically significant. Similar to the analysis in Sec. \ref{subsec:suffix and remaining time prediction}, we employ several paired t-tests.  In particular, for each beam size $n \in \{1,3,5,7,10,15\}$, we pair the corresponding suffix and remaining time predictions provided by two models, i.e., the one trained by MLMME and the one trained using MLE. We consider the following hypotheses for comparing the paired SDL values:
\begin{equation}
    \small{H_0: \mu_d = 0, \quad H_1: \mu_d > 0}
\end{equation}
\noindent Where $\mu_d$ is the average $d = SDL^{(MLMME)} - SDL^{(MLE)}$ between the paired predicted suffixes. Similarly, the following hypotheses are considered for paired  remaining  time absolute  errors:
\begin{equation}
    \small{H_0: \mu_d = 0, \quad H_1: \mu_d < 0}
\end{equation}
\noindent Where $\mu_d$ is the average $d = AE^{(MLMME)} - AE^{(MLE)}$ between the paired remaining time absolute errors.

Table \ref{table: p-value for mlmme and mle} shows the results of applying paired t-tests.  One sees that the superiority of the model trained using MLMME for suffix and remaining time prediction is statistically significant for large datasets BPI12 and BPI17, for all beam sizes. For BPI12(W) and $n=1$, we did not find statistically significant differences in suffix and remaining time prediction. However, for $n>1$, the improvements obtained by the adversarial training are statistically significant. Similarly, for Helpdesk and $n \geq 5$, the positive effects of adversarial training are statistically significant.

\begin{table}[h]
	\centering 
	\footnotesize{ \begin{tabular}{|p{1.3cm}| p{1.2cm}| p{1.3cm} | p{1.0cm} | p{1.0cm} | p{1.2cm}| p{1.3cm}|p{1.0cm}|p{1.0cm}|}
		\hline
    \multicolumn{1}{|c|}{} & \multicolumn{4}{|c|} {\textbf{Average time per iteration (sec.)}}\\
		\hline \hline
		\textbf{Approach} & Helpdesk & BPI12(W) & BPI12 & BPI17 \\ [1ex]
		\hline
		MLE & 1.48    & 104.32  & 225.79 & 158.41\\
		MLMME  & 4.22   & 282.85  & 671.57 & 418.82\\
		\hline
	\end{tabular} }
		\caption{The average execution time per iteration for the proposed training (MLMME) and standard training (MLE).}
	\label{table:training time}
		\vspace{-4mm}
\end{table}

\subsubsection{Time Complexity} Table \ref{table:training time} shows the average execution time per iteration to train the proposed encoder-decoder architecture, see Fig. \ref{fig:encoder-decoder}, using the presented adversarial training method, i.e., MLMME, and the standard training method, i.e., MLE. These measurements are obtained by averaging the recorded execution time in each iteration of training. Although the MLMME method provides more accurate predictions than MLE, one sees it is slower up to 3 times mostly due to the extra updates that the discriminator needs in each iteration of Alg. 1.

\section{Conclusion}
\label{sec: conclusion}
\noindent This paper put forward a novel open-loop adversarial training and an encoder-decoder architecture for the suffix and the remaining time prediction to the realm of sequential temporal data. We adopted the Beam search on top of our architecture to provide the $n$ most probable suffix predictions for a prefix of events, which can be used for further analysis. Comprehensive experiments with statistical tests on four real-world datasets against three baselines show that: i) the approach learns temporal dynamics of sequences more accurately than the baselines; and ii) the proposed training method MLMME boosts a model's performance to capture the temporal relationships of events in sequential data. Deep models are vulnerable to adversarial attacks. Thus, as future work, we plan to build more robust predictive models. Another avenue is to explore other applications 
beyond business process management.
 
\smallskip\noindent\textbf{Acknowledgments} This research is funded by the Australian Research Council (DP180102839). We appreciate the reviewers comments towards improving our manuscript. 

\vspace{-\baselineskip}
\bibliographystyle{siam}
\bibliography{mybibfile}

\end{document}